\documentclass[sigconf]{acmart}

\usepackage{booktabs} 
\usepackage[utf8]{inputenc}
\usepackage{array}
\usepackage{makecell}
\usepackage{algorithm}
\usepackage{algorithmic}
\usepackage{mathtools}
\DeclarePairedDelimiter{\ceil}{\lceil}{\rceil}

\setcopyright{rightsretained}

\acmDOI{10.475/123_4}

\acmConference[ICDSC'18]{ACM International Conference on Distributed Smart Cameras}{September 2018}{Eindhoven, Netherlands}
\acmYear{2018}
\copyrightyear{2018}

\begin{document}
\title{An accurate retrieval through R-MAC+ descriptors for landmark recognition}

\author{Federico Magliani}
\orcid{0000-0001-5526-0449}
\affiliation{%
  \institution{IMP lab - D.I.A. \\Università di Parma}
  \streetaddress{Parco Area delle Scienze, 181/A}
  \city{Parma}
  \state{Italy}
  \postcode{43124}
}
\email{federico.magliani@studenti.unipr.it}

\author{Andrea Prati}
\affiliation{%
  \institution{IMP lab - D.I.A. \\Università di Parma}
  \streetaddress{Parco Area delle Scienze, 181/A}
  \city{Parma}
  \state{Italy}
  \postcode{43124}
}
\email{andrea.prati@unipr.it}

\renewcommand{\shortauthors}{F. Magliani et al.}

\begin{abstract}
The landmark recognition problem is far from being solved, but with the use of features extracted from intermediate layers of Convolutional Neural Networks (CNNs), excellent results have been obtained.
In this work, we propose some improvements on the creation of R-MAC descriptors in order to make the newly-proposed R-MAC+ descriptors more representative than the previous ones.
However, the main contribution of this paper is a novel retrieval technique, that exploits the fine representativeness of the MAC descriptors of the database images. Using this descriptors called "db regions" during the retrieval stage, the performance is greatly improved. 
The proposed method is tested on different public datasets: Oxford5k, Paris6k and Holidays. It outperforms the state-of-the-art results on Holidays and reached excellent results on Oxford5k and Paris6k, overcame only by approaches based on fine-tuning strategies.
\end{abstract}

%
%


\keywords{Landmark Recognition, Content-Based Image Retrieval, Deep Learning}

\maketitle

\section{Introduction}

The landmark recognition problem is among the first tasks treated in computer vision. It consists in the retrieval of the building/place represented in a picture. It may seen simple, but it presentes many difficulties. The changes in viewpoint, illumination conditions, resolution of images and the presence of distractors make this task very interesting to solve.
Obviously, the main objective of this problem is to obtain a high accuracy with an embedding of reduced size in a small amount of time.
The last two constraints are even more important if the retrieval is executed on mobile devices, because the users do not want to wait long time for obtaining the results and the mobile devices usually do not have a large memory space.
In addition, it is important to remind the problem of semantic gap: for a human, this task is pretty simple thanks to personal knowledge and experience, but, for a computer, it is hard because it can use only the information available in the images. 

Currently, with Deep Learning (DL) techniques, the new methods implemented for the solution of the image retrieval problem reached excellent results. The idea behind the use of DL is based on trasferring knowledge gained on classification in retrieval without a significant extra cost. 

The principal advantages of using Convolutional Neural Networks (CNNs) is the possibility to extract features from pre-trained networks. It results to be convenient because is very expensive to train a CNN from scratch and it can be difficult to tune the hyperparameters of the network. Furthermore, the features extracted from pre-trained networks are more representative and it is possible to construct very discriminative embeddings. This is due to the training on large datasets (e.g. ImageNet \cite{deng2009imagenet}) and to the architectures of CNNs that allow to learn high-level features. 
Also with the actual GPUs, it is possible to extract features faster than the old hand-crafted methods (like SIFT \cite{lowe2004distinctive} and SURF \cite{bay2006surf}) \cite{zheng2017sift}. 

A drawback of these new approaches is that the feature extraction phase is executed in a dense way. Recently, in order to overcome this weakness, researchers implemented methods to detect the most discriminative parts or regions in an image.
The rigid-grid mechanism is the most used approach, because it is fast and simple to implement. 
There are some other approaches such as attention methods \cite{jimenez2017class} and  Region Proposal Networks (RPNs) \cite{ren2015faster}. The first ones detected patches from the most active regions obtained from filters of CNNs. Instead, the second ones need training and annotated images, but in most cases they obtain slightly better results than all the other approaches \cite{gordo2016deep}. This is due to a huge number of regions detected, although with a great amount of time spent for training the RPNs.


In this paper, we propose several contributions:
\begin{itemize}
\item a new region detector implemented through grids, that respect the aspect ratio of the image; 
\item an improvement on the effectiveness of the multi-resolution approach for R-MAC descriptors;
\item a novel retrieval method for checking the similarities between query descriptors and regions of database R-MAC descriptors. It allows to outperform the results of R-MAC descriptors on Oxford5k and Paris6k by +7\% and +3\%.
\end{itemize}

This paper is organized as follows. Section \ref{related_works} explains the approaches used in the state of the art. Section \ref{r-mac} briefly reviews the R-MAC descriptors and then describes the contributions of this paper. Then, Section \ref{results} evaluates the proposed methods on public benchmarks: Oxford5k, Paris6k and Holidays. Finally, concluding remarks are reported.

\section{Related works}\label{related_works}


Bag of Words (BoW) \cite{sivic2003video} was the first method implemented for solving image recognition problem. It is based on a vocabulary of the most representative features and every image is described by a vector of occurences of the vocabulary features. Recently, BLCF \cite{mohedano2016bags} used convolutional features in BoW scheme, obtaining good results in retrieval thanks to the sparsity and scalability of the used embedding model.

Another famous classical embedding used in image retrieval is Vector of Locally Aggregated Descriptors (VLAD) \cite{jegou2010aggregating}. It is similar to BoW model, but using the residual of the descriptors, calculated as difference of the feature descriptor and the closest centers in the vocabulary. There are many variant of this embedding, for example: gVLAD \cite{wang2014geometric}, CEVLAD \cite{zhou2016distribution} and locVLAD \cite{magliani2017location}.
Ng \textit{et al.} \cite{ng2015exploiting} used the features of intermediate layers of a CNN to construct VLAD embeddings. Also, Arandjeloic \textit{et al.} \cite{arandjelovic2016netvlad} implemented VLAD descriptors through CNN. They outperformed the results obtained with VLAD using the hand-crafted methods as feature extractor.

In contrast to the previous methods, with the rise of DL for the image retrieval task, many other embeddings based on CNN features have been constructed. 
The important choice related to the global descriptor is the type of pooling (sum, max, VLAD or a combination of them) adopted for the transformation of the feature maps in descriptors. 

Yan \textit{et al.} \cite{yan2016cnn} proposed a three-level representation with the application of max-pooling, sum-pooling and VLAD-pooling.
In contrast to the previous method, that applied the pooling operation on the entire image, Mopuri \textit{et al.} \cite{mopuri2015object} introduced an algorithm of object proposals and then pooled only the features of the detected regions. 

Babenko and Lemptisky \cite{babenko2015aggregating} implemented sum-pooling and showed that this outperforms max-pooling when applied after a process of PCA-whitening.
Tolias \textit{et al.} \cite{tolias2015particular} adopted max-pooling and created Regional Maximum Activations of Convolutions (R-MAC) descriptors, that actually are the most used descriptors in image retrieval problems. 
Many variants of these descriptors have been proposed, in order to overcome their weaknesses. First, Gordo \textit{et al.} \cite{gordo2017end} proposed: a multi-resolution version, that makes the descriptor robust to the scale; a fine-tuning strategy to obtain a representation more specific to the domain of the problem; a RPN for the detection of the relevant regions. All these contributions highly boost the performance, reaching the state-of-the-art results on many public datasets.
Second, Seddati \textit{et al.} \cite{seddati2017towards} implemented a different multi-resolution approach, a new feature selection strategy not based on PCA and a final localization method based on Class Activation Map (CAM) \cite{zhou2016learning}, reaching the state-of-the-art result on Holidays dataset.
Finally, Laskar and Kannala \cite{laskar2017context} introduced a saliency method for a correct weighting of the regions used for the creation of R-MACs. Actually, their performance represent the state of the art on Oxford5k and Paris6k datasets.

\section{R-MAC descriptors} \label{r-mac}

\begin{figure}
\includegraphics[width=4cm]{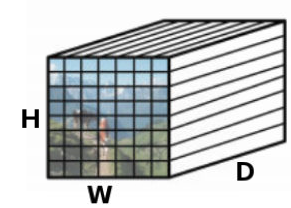}
\caption{Representation of feature maps extracted from a  certain layer of CNN} \label{feature_maps}
\end{figure}

The MAC descriptor \cite{tolias2015particular} of an image $I$ of size $W_I \times H_I$ is created as follows. At the beginning, the layer from which the features need to be extracted is selected, then, as a consequence a 3D tensor of $W \times H \times D$  dimensions is obtained, as it can see in Fig. \ref{feature_maps}, where:
\begin{itemize}
\item $D$ refers to the number of output feature channels;
\item $W$ indicates the width of the feature maps;
\item $H$ indicates the height of the feature maps.
\end{itemize} 

The 3D tensor can be seen as a set of 2D features channel responses $X = \{X_i\}$, $i = \{1, \dots, D\}$ where $X_i$ is the 2D tensor representing the responses of the $i^{th}$ feature channel over the set $\Omega$ of spatial locations, and $X_i(p)$ is the response at a particular position $p$. Finally, the MAC descriptor is obtained through the max-pooling of the $W \times H$ values on each channel:
$$f_{\Omega} = (f_{\Omega,1}, \dots f_{\Omega,i}, \dots f_{\Omega,D})^T \quad with \quad f_{\Omega,i} = \max_{p \in \Omega} X_i(p)$$
At the end, a $L_2$ normalization function is applied on each channel.

Unfortunately, this representation does not encode the location of the activations. 
For this reason, the same authors of MAC descriptors proposed Regional-MAC (R-MAC) descriptors \cite{tolias2015particular}.
The idea behind this approach is to create many MAC descriptors of different regions detected through a rigid-grid mechanism applied on the feature maps. This strategy improves the performance since it increases the importance of features related to small part of the image (local maxima). 
The regions are detected through a square grid of variable dimensions applied at $L$ different scales:
\begin{itemize}
\item at the largest scale ($l=1$), the region size is determined to be as large as possible, so the height and width of the square are equals to $\min(W,H)$;
\item at every other scale ($l=2, \dots, L$), the $l \cdot (l+m-1)$ regions, with usually $m=2$, have width and height equals to $2 \cdot \min(W,H)/(l+1)$.
\end{itemize}

At the end of this process, the MAC descriptors are post-processed with $L_2$ normalization, PCA-whitening and again $L_2$ normalization. 
Finally, the MAC descriptors of each image are sum-pooled and once more $L_2$ normalized, obtaining the final R-MAC descriptor.

Gordo \textit{et al.} \cite{gordo2017end} proposed a multi-resolution approach in order to make the descriptor more robust to the scale.
This strategy consists in the creation of different R-MAC descriptors of the image feeded in input to the CNN. The input image is resized to 3 different input sizes: 550px, 800px and 1050px on the largest size, by retaining the aspect ratio of the image. Then, the 3 obtained R-MAC descriptors are sum-pooled and $L_2$ normalized.

\subsection{Improvements on R-MAC pipeline: R-MAC+ descriptor}

\begin{table*}[ht!]
 \centering
    \begin{tabular}{|c|c|c|c|c|c|c}
    \hline
    \textbf{Symbol} & \textbf{Definition}  &  \textbf{l=0}   & \textbf{l=1}  & \textbf{l=2}  & \textbf{l=3} \\ \hline
   $W$ & width of the feature maps & \multicolumn{4}{|c|}{-} \\ \hline
   $H$ &  height of the feature maps & \multicolumn{4}{|c|}{-} \\ \hline
   $xRegions$ & regions needed to cover the image along the x axis & 1 & 1 or 2 & 3 & 2 \\ \hline
   $yRegions$ & regions needed to cover the image along the y axis & 0 & 1 or 2 & 2 & 3 \\ \hline
   $widthRegion$ & width of the region & $W$ & $\min(W,H)$ & \multicolumn{2}{|c|}{ $\ceil{\frac{2}{l+1} \cdot \min(W,H)}$} \\ \hline  
   $heightRegion$ & height of the region & $H$ & $\min(W,H)$ & \multicolumn{2}{|c|}{ $\ceil{\frac{2}{l+1} \cdot \min(W,H)}$} \\ \hline
   $\Delta W$ & stride of the next region on the horizontal axis &  \multicolumn{4}{|c|}{$W/xRegions$} \\ \hline  
   $\Delta H$ & stride of the next region on the vertical axis & \multicolumn{4}{|c|}{$H/yRegions$} \\ \hline
	\end{tabular}
\caption{Summary of notation and explanation of region detector algorithm.}
\label{paramsSymbol}
\end{table*}

Our pipeline proposes some improvements on the structure of the R-MAC descriptors implemented by Gordo \textit{et al.} \cite{gordo2017end}. 

First, in the multi-resolution approach, we propose to construct R-MAC descriptors obtained through images resized of: 0\%, +25\% and -25\% on the largest size, by retaining the aspect ratio of the images. This strategy should allow to augment the dimensions of the feature maps in order to have more features and therefore local maxima than the previous multi-resolution R-MAC. This approach is connected to the new region detector, that detects a reduced number of regions (15) instead of the 20 of the original one proposed by Tolias \textit{et al.} \cite{tolias2015particular}. 

\begin{figure*}
\includegraphics[width=2.5cm]{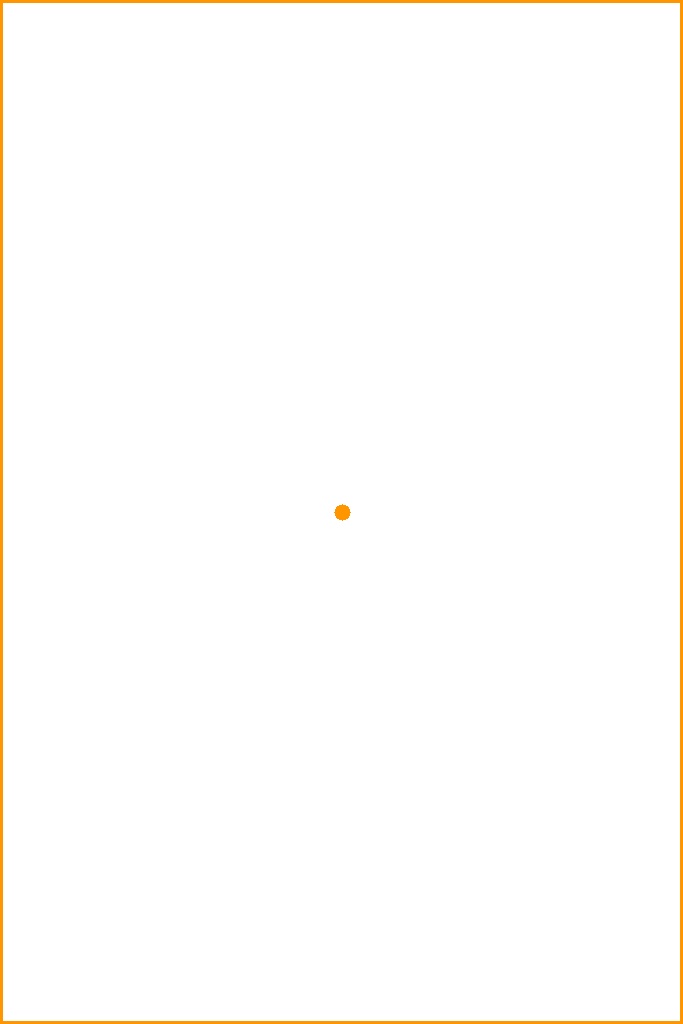}
\includegraphics[width=2.5cm]{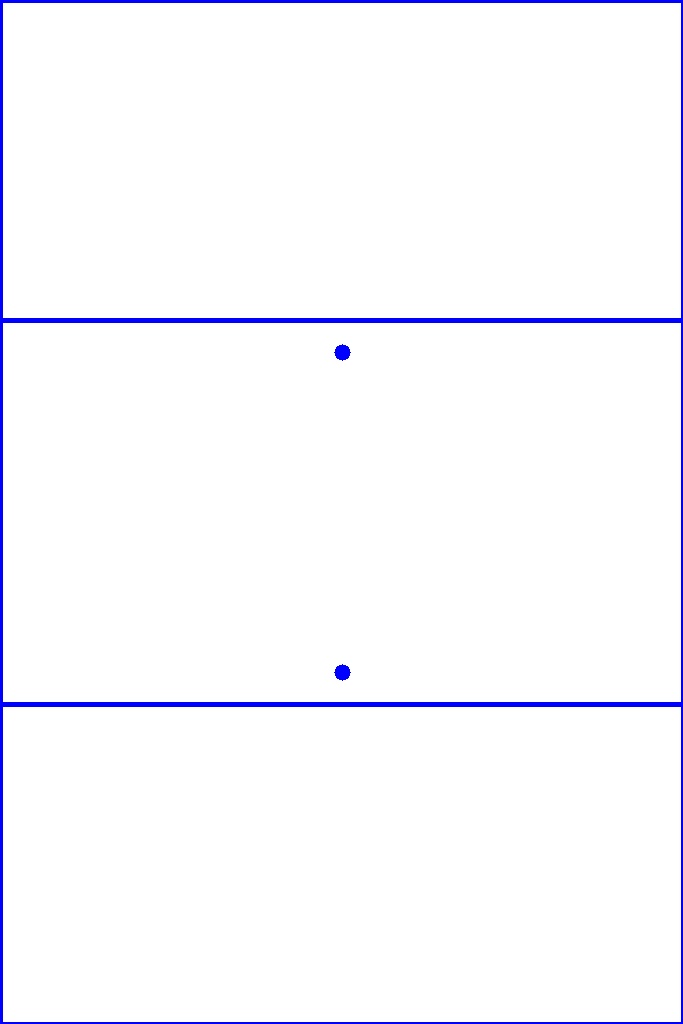}
\includegraphics[width=2.5cm]{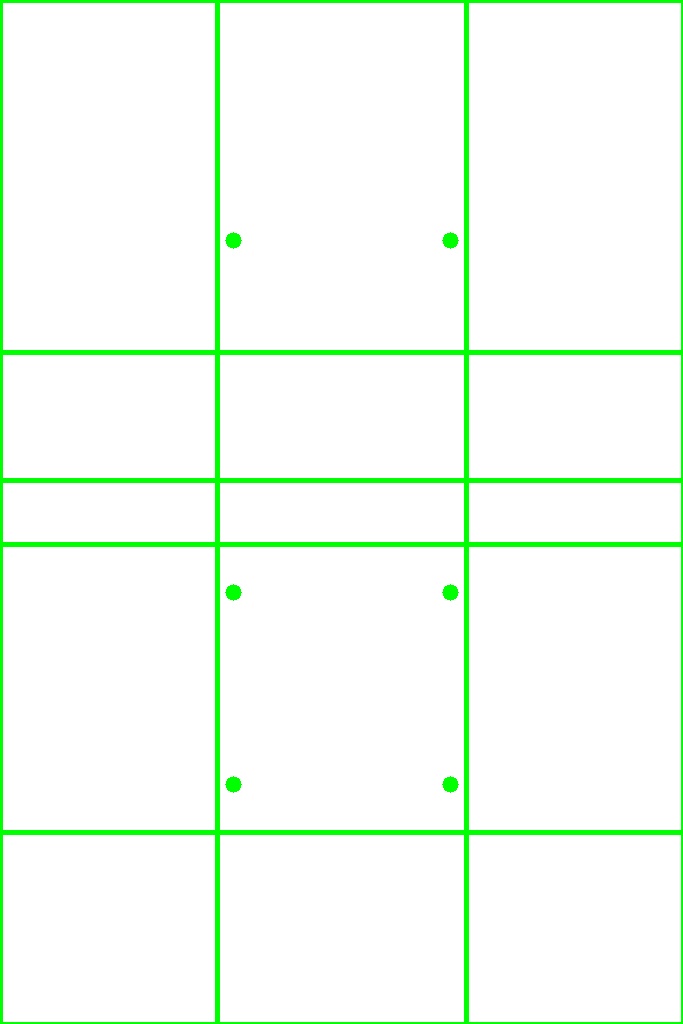}
\includegraphics[width=2.5cm]{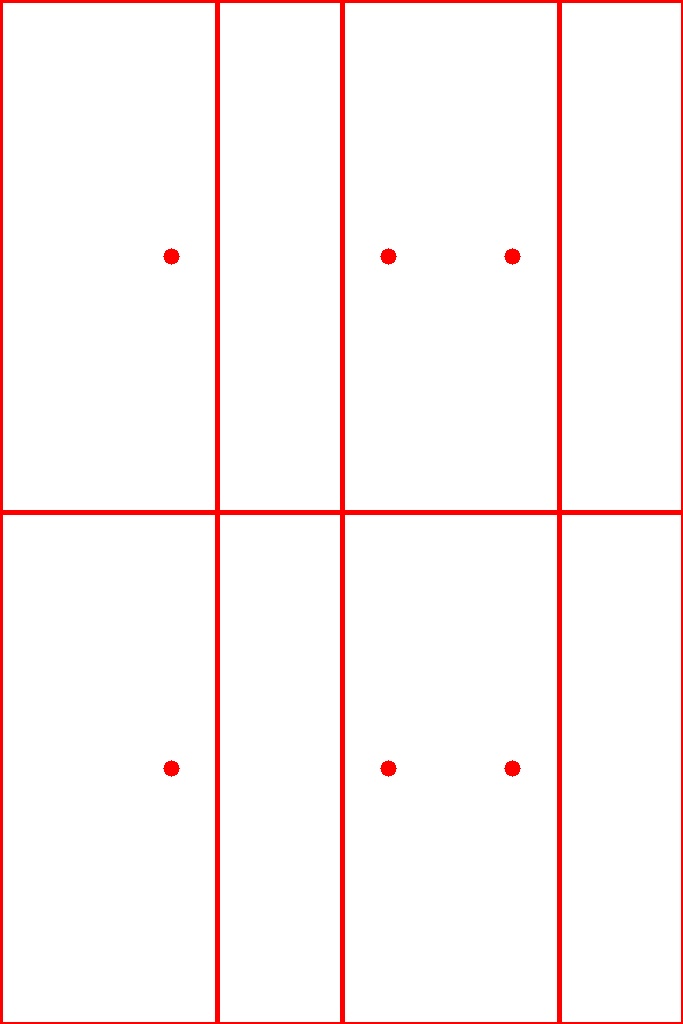}
\caption{Grids adopted in the application of region detector. The circles represent the center of the detected regions. In the first case ($l=0$), 1 orange region is detected, then when $l=1$, 2 blue regions are detected, in the second case when $l=2$, 6 green regions are detected and finally in the last step ($l=3$), 6 red regions are detected. } \label{example_grid_detector}
\end{figure*}

Second, a new grid mechanism for the detection of the regions is proposed. 
The new region detector is structured on $L=3$ different scales. Symbols used and their values are explained in Table \ref{paramsSymbol} and an example is represented in Fig. \ref{example_grid_detector}.
The first level ($l=0$) of the region detector is represented by an orange region covering all the image, that of course has the sizes of the image. Then, in the second level ($l=1$) 2 blue square regions are used, arranged along the largest size of the image. The number of $xRegions$ and $yRegions$, in the case of $l=1$, are determined in the Equations \ref{xRegions_l1} and \ref{yRegions_l1}.


\begin{equation}
\resizebox{.25 \textwidth}{!} 
{$
xRegions  =
  \begin{cases}
   1  & \quad \text{if W < H}\\
    2  & \quad \text{otherwise}
   \end{cases}
$}
\label{xRegions_l1}
\end{equation}

\begin{equation}
\resizebox{.25 \textwidth}{!} 
{$
yRegions  =
  \begin{cases}
   2  & \quad \text{if W < H}\\
    1  & \quad \text{otherwise}
   \end{cases}
$}
\label{yRegions_l1}
\end{equation}


In the third step ($l=2$), 6 green rectangualrs are adopted in order to cover entirely the image. Finally, in the last step ($l=3$), 6 red rectangulares are used.
The value of $heightRegion$ and $widthRegion$ are equal to $\ceil{\frac{2}{l+1} \cdot \min(W,H)}$, but if the regions do not cover entirely the image, the width and the height of the regions change. These values are calculated following the Equations \ref{heightRegion} and \ref{weightRegion}.

\begin{equation}
\resizebox{.45 \textwidth}{!} 
{$
heightRegion  =
  \begin{cases}
   \ceil{H/yRegions}  & \quad \text{if $heightRegion \cdot yRegions < H$}\\
    heightRegion  & \quad \text{otherwise}
   \end{cases}
$}
\label{heightRegion}
\end{equation}

\begin{equation}
\resizebox{.45 \textwidth}{!} 
{$
weightRegion  =
  \begin{cases}
   \ceil{W/xRegions}  & \quad \text{if $weightRegion \cdot xRegions < W$}\\
    weightRegion  & \quad \text{otherwise}
   \end{cases}
$}
\label{weightRegion}
\end{equation}


The regions are arranged based on $\Delta W$ along the horizontal axis and on $\Delta H$ along the vertical axis.
This means that the image will be entirely covered because the $\Delta W$ and $\Delta H$ are related to the dimensions ($W, H$) and to the regions ($xRegions, yRegions$).

It is preferrable to overlap the regions instead of avoiding to cover entirely the image, because otherwise usefull information would be lost.


The grids used in the proposed region detector are squares and rectangulars of different sizes,  as explained in the Table \ref{paramsSymbol}.
As a result, this method produces a lower number of regions than the original one, allowing to reduce the time spent in the creation of R-MAC+ descriptors, without loss in the final accuracy.

\begin{figure}
\includegraphics[width=8cm]{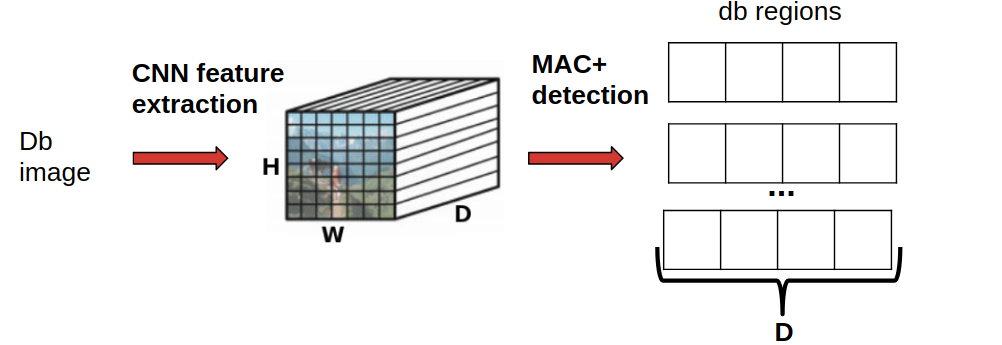}
\caption{Pipeline of creation of "db regions" for the db images} \label{dbImage}
\end{figure}

\begin{figure*}
\includegraphics[width=11cm]{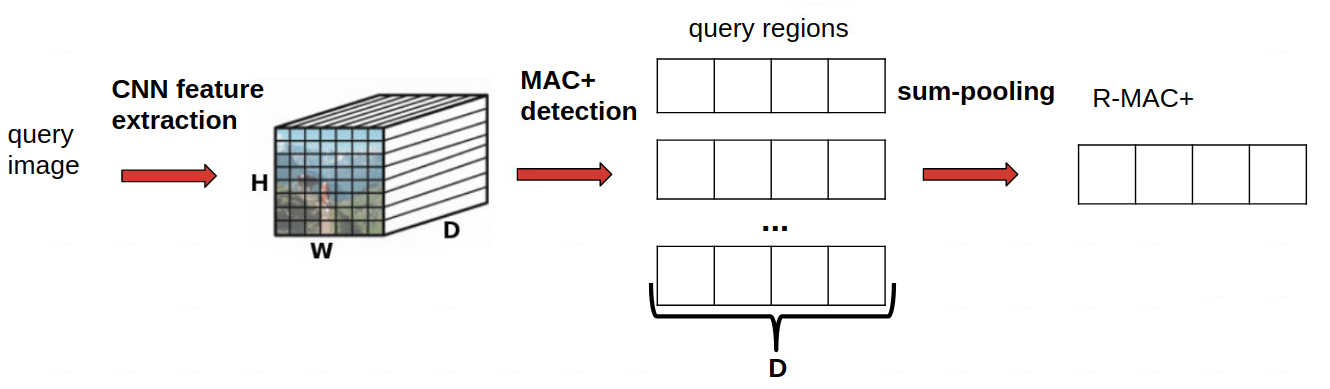}
\caption{Pipeline of creation of R-MAC descriptors for the query images} \label{queryImage}
\end{figure*}

The third improvement introduced is a novel retrieval method. Usually, after the creation of all R-MAC descriptors, for each query image a similarity ranking is constructed through the sorting of database images based on their $L_2$ distances from the query descriptor.
The new retrieval method proposes to modify the descriptors evaluated in the retrieval phase. For the query images it proposes to use R-MAC+ descriptors and for the database images "db regions" are suggested. The "db regions", represented in Fig. \ref{dbImage}, are the MAC+ descriptors obtained during the process of creation of R-MAC+ descriptors of the database images. They represented the maxima of the different regions detected with the grid mechanism. Of course, they are related only to the database images.
The "db regions" are used in the retrieval phase instead of R-MAC+ descriptors of database images because a small part or a region of the image can well represent the entire image. This is due to the features extracted from CNN and max-pooling. Besides, sum-pooling and $L_2$ normalization reduce the value of features of the regions, but this does not happen with the proposed retrieval strategy. 
Instead, on the query side, R-MAC+ descriptors are used as represented in Fig. \ref{queryImage}, because it is preferrable to have an unique query descriptor for the retrieval phase, otherwise if the "query regions" are used there would be confusion or mismatching. 
The advantage of this approach consists in comparing different "db regions" of the database images with the descriptor of the query image, choosing the one that obtains the best similarity or the minimum $L_2$ distance to the query descriptor. 
If the multi-resolution approach is adopted, there will be 45 regions for each database image to check for the retrieval, otherwise only 15 regions. 
This requires more time, but with an important extra boost on the final accuracy retrieval performance.

\section{Experimental results} \label{results}

In order to evaluate the accuracy of the proposed embedding technique with respect to the state of the art, we run experiments on public datasets and employing standard evaluation metrics.

\subsection{Datasets and metrics} \label{datasets}

There are many different image datasets for landmark recognition that are used in order to evaluate the algorithms. The most used are the following:
\begin{itemize}
\item \textbf{Oxford5k} \cite{Philbin07} is composed by 5063 images representing the buildings and the places of Oxford (UK), subdivided in 11 classes. All the images are used as database images and the query images are 55, which are cropped for making more difficult the querying phase;
\item \textbf{Paris6k} \cite{Philbin08} is composed by 6412 images representing the buildings and the places of Paris (France), subdivided in 12 classes. All the images are used as database images and the query images are 55, which are cropped for making more difficult the querying phase;
\item \textbf{Holidays} \cite{jegou2008hamming} is composed of 1491 high-resolution images representing the holidays photos of different locations and objects, subdivided in 500 classes. The database images are 991 and the query images are 500, one for every class. 
\end{itemize}

According to \cite{laskar2017context}, on Oxford5k and Paris6k, the query images feed in input to the CNN are the full images, but the R-MAC descriptors are not calculated on the entire feature maps. The descriptors are calculated on the cropped activations, obtained through the projection of bounding box of the query images along the CNN stages. This strategy slightly improves the final accuray results.

To evaluate the accuracy in the retrieval phase, mean Average Precision (mAP) is used.

In order to compare a query image with the database, $L_2$ distance is employed. 

All the experiments are executed on GeForce GTX 1070. For the feature extraction step, Keras python library is used.

\subsection{Results on Oxford5k, Paris6k and Holidays} \label{results_holidays}

Table \ref{table_results_holidays} reports our results obtained on: Oxford5k, Paris6k and Holidays.

\begin{table}[htb]
\centering
    \begin{tabular}{|c|c|c|c|c|}
    \hline
     \footnotesize \textbf{Method} & \footnotesize \textbf{CNN} &  \scriptsize \textbf{Oxford5K} & \scriptsize \textbf{Paris6k} & \scriptsize \textbf{Holidays}  \\ \hline
    MAC$^\dagger$ & VGG19    & 57.44 & 73.15 & 76.26  \\ \hline 
    R-MAC$^\dagger$  & VGG19   & 65.56 & 82.80 & 87.65  \\ \hline 
    R-MAC$^\dagger$  & ResNet50  & 71.77 & 83.31 & 92.55   \\ \hline 
    M-R R-MAC+ & ResNet50  & 78.88 & 88.63 & \textbf{94.63}/95.58 \\ \hline   
    \makecell{M-R R-MAC+ \\ with retrieval based \\ on "db regions"}  & ResNet50 & \textbf{85.39} & \textbf{91.90} & 94.37/\textbf{95.87} \\ \hline


    \end{tabular}
\caption{Results of the proposed methods obtained on some public datasets. $^\dagger$  indicates that the method is re-implemented. M-R indicates that the multi-resolution approach is adopted. The results after / on Holidays represent the experiments executed on the rotated version of the dataset.}
\label{table_results_holidays}
\end{table}

In the first two experiments, VGG19 is adopted for the feature extraction step. It allows to extract feature maps of 512D, instead through ResNet50 is possible to extract feature maps of 2048D, reaching more accurate results than the first experiments thanks to the depth of the CNN architecture.
The features are extracted from the layer "block5\_ pool" on VGG19 and from the layer "activation\_43" on ResNet50.

\begin{table}[hbt]
\centering
    \begin{tabular}{|c|c|c|c|c|}
    \hline
     \small \textbf{Method} & \small \textbf{CNN} &  \scriptsize \textbf{Oxford5K} & \scriptsize \textbf{Paris6k} & \scriptsize \textbf{Holidays}  \\ \hline
   \makecell{M-R R-MAC+ \\ and query expansion} & ResNet50  & 86.45 & 92.01 & \textbf{94.97}/95.97 \\ \hline   
    \makecell{M-R R-MAC+ \\ with retrieval based \\ on "db regions" \\ and query expansion}  & ResNet50 & 87.92 & \textbf{93.64} & 94.42/\textbf{96.05} \\ \hline
 \makecell{M-R R-MAC+ \\ with retrieval based \\ on "db regions" \\ and query expansion \\ of "db regions"}  & ResNet50 & \textbf{88.78} & 92.30 & 94.28/95.91 \\ \hline
     \end{tabular}
\caption{Results of the proposed methods with query expansion techniques obtained on some public datasets. $^\dagger$  indicates that the method is re-implemented. M-R indicates that the multi-resolution approach is adopted. The results after / on Holidays represent the experiments executed on the rotated version of the dataset.}
\label{table_results_holidays_qe}
\end{table}

In Table \ref{table_results_holidays} are presented the experiments executed without any re-ranking strategies, while in Table \ref{table_results_holidays_qe} reports results obtained using query expansion methods in order to improve the final accuracy.
M-R RMAC+ with the retrieval based on "db regions" reached the best results on all the 3 public datasets. Moreover, in the cases of Oxford5k and Paris6k, the improvements compared to the previous experiments are remarkable. Unfortunately, the average query time on Oxford5k and Paris6k increases from 0.04s and 0.11s to 1.25s and 1.54s because a brute-force approach has been applied and the number of descriptors to be compared have been increased by 45 times.

For the query expansion approach, it has been decided to expand the query with a fixed number of descriptors, as explained in \cite{chum2007total}. After some tests, the numbers of top-ranked desriptors used for the query expansion are: 8 for Oxford5k, 6 for Paris6k and 1 for Holidays. This is due to the number of images belonging to the classes: in Holidays, the majority of the classes are composed by a reduced number of elements, instead in Oxford5k and Paris6k, the classes are constituted by many images.
Furthermore, the query expansion of "db regions" produces improvements only on Oxford5k. This method uses for the query expansion the "db regions" instead of the R-MAC+ descriptors of the database images. 

\subsection{Comparison with the state of the art} \label{comparison}

In order to have a fair comparison, Table \ref{table_comparison} reports our results with those of several other state-of-the-art methods on the same datasets.

\begin{table}[hbt]
\centering
    \begin{tabular}{|c|c|c|c|c|}
    \hline
     \small \textbf{Method} &  \small \textbf{Dim.} &  \scriptsize \textbf{Oxford5K} & \scriptsize \textbf{Paris6k} & \scriptsize \textbf{Holidays}  \\ \hline
    VLAD \cite{jegou2010aggregating} & 4096 & 37.80 & 38.60 & 55.60  \\ \hline
    gVLAD \cite{wang2014geometric}  & 128 & 60.00 & - & 77.90   \\ \hline
    NetVLAD \cite{arandjelovic2016netvlad}  & 4096 & 71.60 & 79.70 & 81.70   \\ \hline
    Ng \textit{et al.} \cite{ng2015exploiting}  & 128 & 55.80* & 58.30* & 83.60   \\ \hline 
    Neural codes \cite{babenko2014neural} & 128 & 55.70* & - & 78.90   \\ \hline 
    Babenko \textit{et al.} \cite{babenko2015aggregating} & 256 & 65.70 & - & 78.40 \\ \hline
    Kalantidis \textit{et al.} \cite{kalantidis2016cross} & 512 & 68.20 & 79.70 & 83.10  \\ \hline
    Yan \textit{et al.} \cite{yan2016cnn}  & 128 & - & 76.76 & 84.13   \\ \hline 
    Mopuri \textit{et al.} \cite{mopuri2015object}  & 128 & - & 70.39 & 85.09  \\ \hline 
    BLCF \cite{mohedano2016bags} & 25k & 73.80 & 82.00 & - \\ \hline
    BLCF-SalGAN \cite{mohedano2017saliency} & 336 & 74.60 & 81.20 & - \\ \hline
    R-MAC \cite{tolias2015particular}  & 512 & 66.90 & 83.00 & 85.20  \\ \hline 
    Gordo \textit{et al.} 2016 \cite{gordo2016deep}  & 512 & 83.10 & 87.10 & 86.70   \\ \hline 
    Gordo \textit{et al.} 2017 \cite{gordo2017end} & 2048 & 86.10 & 94.50 & 90.30/94.48 \\ \hline
    Laskar \textit{et al.} \cite{laskar2017context} & 2048 & \textbf{90.20} & \textbf{95.80} & -  \\ \hline
    Seddati \textit{et al.} \cite{seddati2017towards} & variable & 72.27 & 87.10 & 94.00 \\ \hline
    \makecell{\textbf{R-MAC+} \\ \textbf{with retrieval} \\ \textbf{based on} \\ \textbf{"db regions"}} & 2048 & 85.39 & 91.90 & \textbf{94.37}/\textbf{95.87} \\
\hline
    \end{tabular}
\caption{Comparison of state-of-the-art methods on different datasets. * indicates that the method is applied on the full-size query images. The results after / on Holidays represent the experiments executed on the rotated version of the dataset.}
\label{table_comparison}
\end{table}

As reported in the introduction and in related works, the methods that implemented classical embedding as VLAD based on CNNs \cite{arandjelovic2016netvlad, ng2015exploiting} obtain better results than the ones based on hand-crafted methods for the feature extraction phase \cite{jegou2010aggregating, wang2014geometric}.  

The methods based on sum-pooling \cite{babenko2015aggregating, kalantidis2016cross} obtained worse results than the methods based on max-pooling \cite{tolias2015particular}.

\begin{table}[hbt]
\centering
    \begin{tabular}{|c|c|c|c|c|}
    \hline
     \small \textbf{Method} &  \small \textbf{Dim.} &  \scriptsize \textbf{Oxford5K} & \scriptsize \textbf{Paris6k} & \scriptsize \textbf{Holidays}  \\ \hline
     R-MAC \cite{tolias2015particular}  & 512 & 77.30 & 86.50 & -  \\ \hline 
     Kalantidis \textit{et al.} \cite{kalantidis2016cross} & 512 & 72.20 & 85.50 & -  \\ \hline
	 Azizpour \textit{et al.} \cite{azizpour2016factors} &  4096 & 79.00 & 85.10 & 90.00 \\ \hline
     Gordo \textit{et al.} 2016 \cite{gordo2016deep} & 2048 & 89.00 & 93.80 & - \\ \hline
     Gordo \textit{et al.} 2017 \cite{gordo2017end} & 2048 & \textbf{90.60} & \textbf{96.00} & - \\ \hline
    \makecell{\textbf{R-MAC+} \\ \textbf{with retrieval} \\ \textbf{based on} \\ \textbf{"db regions"}} & 2048 & 87.92 & 93.64 & \textbf{94.42}/\textbf{96.05} \\
     \hline
         \end{tabular}
\caption{Comparison of state-of-the-art methods with query expansion techniques on different datasets. The results after / on Holidays represent the experiments executed on the rotated version of the dataset.}
\label{table_comparison_qe}
\end{table}

Moreover, Gordo \textit{et al.}, in \cite{gordo2016deep, gordo2017end}, with their recent improvements due to fine-tuning raised excellent result in the image retrieval task.
Laskar \textit{et al.} \cite{laskar2017context} improved the R-MAC pipeline through a saliency weighting of the regions extracted.

The proposed method R-MAC+ with retrieval based on "db regions" outperforms the state of the art on Holidays and reached good results on Oxford5k and Paris6k, overcame only by methods based on fine-tuning strategies  \cite{gordo2017end, gordo2016deep, laskar2017context}.
The main drawback of fine-tuning is that, if more data to retrieve than the actuals are added then a new fine-tuning procedure needs to be issued, requiring more time as well as additional labelled data.
With the addition of query expansion technique, the R-MAC of Gordo \textit{et al.} \cite{gordo2016deep, gordo2017end} also outperformed our approach, as reported in Table \ref{table_comparison_qe} because they obtained better results than ours in the first stage of the retrieval problem. Besides, the performance gap between our approach and the R-MAC of Gordo is maintained similar before and after the application of query expansion techniques. However, the proposed approach still shows state-of-the-art mAP on the Holidays dataset and on Oxford5k, Paris6k without the application of the fine-tuning strategy.

\section{Conclusions} \label{conclusions}
In this work we propose different improvements on R-MAC descriptors in order to make the retrieval very accurate. 
The proposed multi-resolution approach improves the performance through the use of bigger feature maps than the initial multi-resolution approach. The new region detector with the use of adaptable grids allows to catch more local maxima, that are the relevant features for the MAC descriptors. 
Finally, the novel retrieval method based on "db regions" highly boosts the performance on Oxford5k and Paris6k.
The proposed method outperforms the state of the art on Holidays, both on the original and rotated version. Also it outperforms the state-of-the-art results on some other public benchmarks without the fine-tuning application.
The proposed strategies (both fot the descriptors and the retrieval methods) are general enough to be applicable to the image retrieval tasks.

\textbf{Acknowledgments}. This work is partially funded by Regione Emilia Romagna under the “Piano triennale alte competenze per la ricerca, il trasferimento tecnologico e l’imprenditorialita”.


\bibliographystyle{ACM-Reference-Format}
\bibliography{sample-bibliography}

\end{document}